\PassOptionsToPackage{unicode}{hyperref}
\PassOptionsToPackage{hyphens}{url}
\PassOptionsToPackage{dvipsnames,svgnames,x11names}{xcolor}
\documentclass[
]{article}

\usepackage{amsmath,amssymb}
\usepackage{iftex}
\ifPDFTeX
  \usepackage[T1]{fontenc}
  \usepackage[utf8]{inputenc}
  \usepackage{textcomp} 
\else 
  \usepackage{unicode-math}
  \defaultfontfeatures{Scale=MatchLowercase}
  \defaultfontfeatures[\rmfamily]{Ligatures=TeX,Scale=1}
\fi
\usepackage{lmodern}
\ifPDFTeX\else  
\fi
\IfFileExists{upquote.sty}{\usepackage{upquote}}{}
\IfFileExists{microtype.sty}{
  \usepackage[]{microtype}
  \UseMicrotypeSet[protrusion]{basicmath} 
}{}
\makeatletter
\@ifundefined{KOMAClassName}{
  \IfFileExists{parskip.sty}{%
    \usepackage{parskip}
  }{
    \setlength{\parindent}{0pt}
    \setlength{\parskip}{6pt plus 2pt minus 1pt}}
}{
  \KOMAoptions{parskip=half}}
\makeatother
\usepackage{xcolor}
\ifx\paragraph\undefined\else
  \let\oldparagraph\paragraph
  \renewcommand{\paragraph}[1]{\oldparagraph{#1}\mbox{}}
\fi
\ifx\subparagraph\undefined\else
  \let\oldsubparagraph\subparagraph
  \renewcommand{\subparagraph}[1]{\oldsubparagraph{#1}\mbox{}}
\fi

\usepackage{longtable,booktabs,array}
\usepackage{calc} 
\usepackage{etoolbox}
\makeatletter
\patchcmd\longtable{\par}{\if@noskipsec\mbox{}\fi\par}{}{}
\makeatother
\IfFileExists{footnotehyper.sty}{\usepackage{footnotehyper}}{\usepackage{footnote}}
\makesavenoteenv{longtable}
\usepackage{graphicx}
\makeatletter
\def\maxwidth{\ifdim\Gin@nat@width>\linewidth\linewidth\else\Gin@nat@width\fi}
\def\maxheight{\ifdim\Gin@nat@height>\textheight\textheight\else\Gin@nat@height\fi}
\makeatother
\setkeys{Gin}{width=\maxwidth,height=\maxheight,keepaspectratio}
\makeatletter
\def\fps@figure{htbp}
\makeatother
\newlength{\cslhangindent}
\newlength{\csllabelwidth}
\newlength{\cslentryspacingunit} 
\newenvironment{CSLReferences}[2] 
 {
  \setlength{\parindent}{0pt}
  \ifodd #1
  \let\oldpar\par
  \def\par{\hangindent=\cslhangindent\oldpar}
  \fi
  \setlength{\parskip}{#2\cslentryspacingunit}
 }%
 {}
\usepackage{calc}

\usepackage{arxiv}
\usepackage{orcidlink}
\usepackage{amsmath}
\usepackage[T1]{fontenc}
\makeatletter
\@ifpackageloaded{caption}{}{\usepackage{caption}}
\AtBeginDocument{%
\ifdefined\contentsname
  \renewcommand*\contentsname{Table of contents}
\else
  \newcommand\contentsname{Table of contents}
\fi
\ifdefined\listfigurename
  \renewcommand*\listfigurename{List of Figures}
\else
  \newcommand\listfigurename{List of Figures}
\fi
\ifdefined\listtablename
  \renewcommand*\listtablename{List of Tables}
\else
  \newcommand\listtablename{List of Tables}
\fi
\ifdefined\figurename
  \renewcommand*\figurename{Figure}
\else
  \newcommand\figurename{Figure}
\fi
\ifdefined\tablename
  \renewcommand*\tablename{Table}
\else
  \newcommand\tablename{Table}
\fi
}
\@ifpackageloaded{float}{}{\usepackage{float}}
\floatstyle{ruled}
\@ifundefined{c@chapter}{\newfloat{codelisting}{h}{lop}}{\newfloat{codelisting}{h}{lop}[chapter]}
\floatname{codelisting}{Listing}

\makeatother
\makeatletter
\@ifpackageloaded{caption}{}{\usepackage{caption}}
\@ifpackageloaded{subcaption}{}{\usepackage{subcaption}}
\makeatother
\makeatletter
\@ifpackageloaded{tcolorbox}{}{\usepackage[skins,breakable]{tcolorbox}}
\makeatother
\makeatletter
\@ifundefined{shadecolor}{\definecolor{shadecolor}{rgb}{.97, .97, .97}}
\makeatother
\ifLuaTeX
  \usepackage{selnolig}  
\fi
\IfFileExists{bookmark.sty}{\usepackage{bookmark}}{\usepackage{hyperref}}
\IfFileExists{xurl.sty}{\usepackage{xurl}}{} 
\hypersetup{
  pdftitle={Augmenty: A Python Library for Structured Text Augmentation},
  pdfkeywords={Python, natural language
processing, spacy, augmentation},
  colorlinks=true,
  linkcolor={blue},
  filecolor={Maroon},
  citecolor={Blue},
  urlcolor={Blue},
  pdfcreator={LaTeX via pandoc}}

\newcommand{\runninghead}{A Preprint }
\renewcommand{\runninghead}{Augmenty }
\title{Augmenty: A Python Library for Structured Text Augmentation}
\author{\textbf{Kenneth Enevoldsen}~\orcidlink{0000-0001-8733-0966}\\}
\date{}
\begin{document}
\maketitle
\begin{abstract}
Augmnety is a Python library for structured text augmentation. It is
built on top of spaCy and allows for augmentation of both the text and
its annotations. Augmenty provides a wide range of augmenters which can
be combined in a flexible manner to create complex augmentation
pipelines. It also includes a set of primitives that can be used to
create custom augmenters such as word replacement augmenters. This
functionality allows for augmentations within a range of applications
such as named entity recognition (NER), part-of-speech tagging, and
dependency parsing.
\end{abstract}
{\bfseries \emph Keywords}
\def\sep{\textbullet\ }
Python \sep natural language processing \sep spacy \sep 
augmentation

\ifdefined\Shaded\renewenvironment{Shaded}{\begin{tcolorbox}[frame hidden, boxrule=0pt, borderline west={3pt}{0pt}{shadecolor}, interior hidden, enhanced, sharp corners, breakable]}{\end{tcolorbox}}\fi

\hypertarget{summary}{%
\section{Summary}\label{summary}}

Text augmentation is useful for tool for training (Wei and Zou 2019) and
evaluating (Ribeiro et al. 2020) natural language processing models and
systems. Despite its utility existing libraries for text augmentation
often exhibit limitations in terms of functionality and flexibility,
being confined to basic tasks such as text-classification or cater to
specific downstream use-cases such as estimating robustness (Goel et al.
2021). Recognizing these constraints, \texttt{Augmenty} is a tool for
structured text augmentation of the text along with its annotations.
\texttt{Augmenty} integrates seamlessly with the popular NLP library
\texttt{spaCy} (Honnibal et al. 2020) and seeks to be compatible with
all models and tasks supported by \texttt{spaCy}. Augmenty provides a
wide range of augmenters which can be combined in a flexible manner to
create complex augmentation pipelines. It also includes a set of
primitives that can be used to create custom augmenters such as word
replacement augmenters. This functionality allows for augmentations
within a range of applications such as named entity recognition (NER),
part-of-speech tagging, and dependency parsing.

\hypertarget{statement-of-need}{%
\section{Statement of need}\label{statement-of-need}}

Augmentation is a powerful tool within disciplines such as computer
vision (Wang, Perez, et al. 2017) and speech recognition (Park et al.
2019) and it used for both training more robust models and evaluating
models ability to handle pertubations. Within natural language
processing (NLP) augmentation has seen some uses as a tool for
generating additional training data (Wei and Zou 2019), but have really
shined as a tool for model evaluation, such as estimating robustness
(Goel et al. 2021) and bias (Lassen et al. 2023), or for creating novel
datasets (Nielsen 2023).

Despite its utility, existing libraries for text augmentation often
exhibit limitations in terms of functionality and flexibility. Commonly
they only provide pure string augmentation which typically leads to the
annotations becoming misaligned with the text. This has limited the use
of augmentation to tasks such as text classification while neglected
structured prediction tasks such as named entity recognition (NER) or
coreference resolution. This has limited the use of augmentation to a
wide range of tasks both for training and evaluation.

Existing tools such as \texttt{textgenie} (Pandya 2023), and
\texttt{textaugment} (Marivate and Sefara 2020) implements powerful
techniques such as backtranslation and paraprashing, which are useful
for augmentation for text-classification tasks. However, these tools
neglect a category of tasks which require that the annotations are
aligned with the augmentation of the text. For instance even simple
augmentation such as replacing the named entity ``Jane Doe'' with
``John'' will lead to a misalignment of the NER annotation,
part-of-speech tags, etc., which if not properly handled will lead to a
misinterpretation of the model performance or generation of incorrect
training samples.

\texttt{Augmenty} seeks to remedy this by providing a flexible and
easy-to-use interface for structured text augmentation.
\texttt{Augmenty} is built to integrate well with of the \texttt{spaCy}
(Honnibal et al. 2020) and seeks to be compatible with the broads set of
tasks supported by \texttt{spaCy}. Augmenty provides augmenters which
takes in a spaCy \texttt{Doc}-object (but works just as well with
\texttt{string}-objects) and returns a new \texttt{Doc}-object with the
augmentations applied. This allows for augmentations of both the text
and the annotations present in the \texttt{Doc}-object.

Other tools for data augmentation focus on specific downstream
application such \texttt{textattack} (Morris et al. 2020) which is
useful for adversarial attacks of classification systems or
\texttt{robustnessgym} (Goel et al. 2021) which is useful for evaluating
robustness of classification systems. \texttt{Augmenty} does not seek to
replace any of these tools but seeks to provide a general purpose tool
for augmentation of both the text and its annotations. This allows for
augmentations within a range of applications such as named entity
recognition, part-of-speech tagging, and dependency parsing.

\hypertarget{features-functionality}{%
\section{Features \& Functionality}\label{features-functionality}}

\texttt{Augmenty} is a Python library that implements augmentation based
on \texttt{spaCy}'s \texttt{Doc} object. \texttt{spaCy}'s \texttt{Doc}
object is a container for a text and its annotations. This makes it easy
to augment text and annotations simultaneously. The \texttt{Doc} object
can easily be extended to include custom augmention not available in
\texttt{spaCy} by adding custom attributes to the \texttt{Doc} object.
While \texttt{Augmenty} is built to augment \texttt{Doc}s the object is
easily converted into strings, lists or other formats. The annotations
within a \texttt{Doc} can be provided either by existing annotations or
by annotations provided by an existing model.

Augmenty implements a series of augmenters for token-, span- and
sentence-level augmentation. These augmenters range from primitive
augmentations such as word replacement which can be used to quickly
construct new augmenters to language specific augmenters such as
keystroke error augmentations based on a French keyboard layout.
Augmenty also integrates with other libraries such as \texttt{NLTK}
{[}bird2009natural{]} to allow for augmentations based on WordNet
(Miller 1994) and allows for specification of static word vectors
{[}pennington-etal-2014-glove{]} to allow for augmentations based on
word similarity. Lastly, \texttt{augmenty} provides a set of utility
functions for repeating augmentations, combining augmenters or adjust
the percentage of documents that should be augmented. This allow for the
flexible construction of augmentation pipelines specific to the task at
hand.

Augmenty is furthemore designed to be compatible with \texttt{spaCy} and
thus its augmenters can easily be utilized during the training of
\texttt{spaCy} models.

\hypertarget{example-use-cases}{%
\section{Example Use Cases}\label{example-use-cases}}

Augmenty have already seen used in a number of projects. The code base
was initially developed for evaluating the robustness and bias of
\texttt{DaCy} (Enevoldsen, Hansen, and Nielbo 2021), a state-of-the-art
Danish NLP pipeline. It is also continually used to evaluate Danish NER
systems for biases and robustness on the DaCy website. Augmenty has also
been used to detect intersectional biases (Lassen et al. 2023) and used
within benchmark of Danish language models (Sloth and Rybner 2023).

Besides its existing use-cases \texttt{Augmenty} could for example also
be used to a) upsample minority classes without duplicating samples, b)
train less biased models by e.g.~replacing names with names of minority
groups c) train more robust models e.g.~by augmenting with typos or d)
generate pseudo historical data by augmenting with known spelling
variations of words.

\hypertarget{target-audience}{%
\section{Target Audience}\label{target-audience}}

The package is mainly targeted at NLP researchers and practitioners who
wish to augment their data for training or evaluation. The package is
also targeted at researchers who wish to evaluate their models either
augmentations or generating new datasets.

\hypertarget{acknowledgements}{%
\section{Acknowledgements}\label{acknowledgements}}

The authors thank the
\href{https://github.com/KennethEnevoldsen/augmenty/graphs/contributors}{contributors}
of the package notably Lasse Hansen which provided meaningful feedback
on the design of the package at early stages of development.

\hypertarget{refs}{}
\begin{CSLReferences}{1}{0}
\leavevmode\vadjust pre{\hypertarget{ref-Enevoldsen_DaCy_A_Unified_2021}{}}%
Enevoldsen, Kenneth, Lasse Hansen, and Kristoffer L. Nielbo. 2021.
{``{DaCy: A Unified Framework for Danish NLP}.''}
\url{https://ceur-ws.org/Vol-2989/short_paper24.pdf}.

\leavevmode\vadjust pre{\hypertarget{ref-goel-etal-2021-robustness}{}}%
Goel, Karan, Nazneen Fatema Rajani, Jesse Vig, Zachary Taschdjian, Mohit
Bansal, and Christopher Ré. 2021. {``Robustness Gym: Unifying the {NLP}
Evaluation Landscape.''} In \emph{Proceedings of the 2021 Conference of
the North American Chapter of the Association for Computational
Linguistics: Human Language Technologies: Demonstrations}, edited by Avi
Sil and Xi Victoria Lin, 42--55. Online: Association for Computational
Linguistics. \url{https://doi.org/10.18653/v1/2021.naacl-demos.6}.

\leavevmode\vadjust pre{\hypertarget{ref-spacy}{}}%
Honnibal, Matthew, Ines Montani, Sofie Van Landeghem, and Adriane Boyd.
2020. {``{spaCy}: Industrial-Strength Natural Language Processing in
Python.''} \url{https://doi.org/10.5281/zenodo.1212303}.

\leavevmode\vadjust pre{\hypertarget{ref-lassen-etal-2023-detecting}{}}%
Lassen, Ida Marie S., Mina Almasi, Kenneth Enevoldsen, and Ross Deans
Kristensen-McLachlan. 2023. {``Detecting Intersectionality in {NER}
Models: A Data-Driven Approach.''} In \emph{Proceedings of the 7th Joint
SIGHUM Workshop on Computational Linguistics for Cultural Heritage,
Social Sciences, Humanities and Literature}, edited by Stefania
Degaetano-Ortlieb, Anna Kazantseva, Nils Reiter, and Stan Szpakowicz,
116--27. Dubrovnik, Croatia: Association for Computational Linguistics.
\url{https://doi.org/10.18653/v1/2023.latechclfl-1.13}.

\leavevmode\vadjust pre{\hypertarget{ref-marivate2020improving}{}}%
Marivate, Vukosi, and Tshephisho Sefara. 2020. {``Improving Short Text
Classification Through Global Augmentation Methods.''} In
\emph{International Cross-Domain Conference for Machine Learning and
Knowledge Extraction}, 385--99. Springer.

\leavevmode\vadjust pre{\hypertarget{ref-miller-1994-wordnet}{}}%
Miller, George A. 1994. {``{W}ord{N}et: A Lexical Database for
{E}nglish.''} In \emph{{H}uman {L}anguage {T}echnology: Proceedings of a
Workshop Held at {P}lainsboro, {N}ew {J}ersey, {M}arch 8-11, 1994}.
\url{https://aclanthology.org/H94-1111}.

\leavevmode\vadjust pre{\hypertarget{ref-morris2020textattack}{}}%
Morris, John, Eli Lifland, Jin Yong Yoo, Jake Grigsby, Di Jin, and
Yanjun Qi. 2020. {``TextAttack: A Framework for Adversarial Attacks,
Data Augmentation, and Adversarial Training in NLP.''} In
\emph{Proceedings of the 2020 Conference on Empirical Methods in Natural
Language Processing: System Demonstrations}, 119--26.

\leavevmode\vadjust pre{\hypertarget{ref-nielsen-2023-scandeval}{}}%
Nielsen, Dan. 2023. {``{S}cand{E}val: A Benchmark for {S}candinavian
Natural Language Processing.''} In \emph{Proceedings of the 24th Nordic
Conference on Computational Linguistics (NoDaLiDa)}, edited by Tanel
Alumäe and Mark Fishel, 185--201. T{ó}rshavn, Faroe Islands: University
of Tartu Library. \url{https://aclanthology.org/2023.nodalida-1.20}.

\leavevmode\vadjust pre{\hypertarget{ref-pandya_hetpandyatextgenie_2023}{}}%
Pandya, Het. 2023. {``Hetpandya/Textgenie.''}
\url{https://github.com/hetpandya/textgenie}.

\leavevmode\vadjust pre{\hypertarget{ref-Park2019SpecAugmentAS}{}}%
Park, Daniel S., William Chan, Yu Zhang, Chung-Cheng Chiu, Barret Zoph,
Ekin Dogus Cubuk, and Quoc V. Le. 2019. {``SpecAugment: A Simple Data
Augmentation Method for Automatic Speech Recognition.''} In
\emph{Interspeech}.
\url{https://api.semanticscholar.org/CorpusID:121321299}.

\leavevmode\vadjust pre{\hypertarget{ref-ribeiro-etal-2020-beyond}{}}%
Ribeiro, Marco Tulio, Tongshuang Wu, Carlos Guestrin, and Sameer Singh.
2020. {``Beyond Accuracy: Behavioral Testing of {NLP} Models with
{C}heck{L}ist.''} In \emph{Proceedings of the 58th Annual Meeting of the
Association for Computational Linguistics}, edited by Dan Jurafsky,
Joyce Chai, Natalie Schluter, and Joel Tetreault, 4902--12. Online:
Association for Computational Linguistics.
\url{https://doi.org/10.18653/v1/2020.acl-main.442}.

\leavevmode\vadjust pre{\hypertarget{ref-sloth_dadebiasgenda-lens_2023}{}}%
Sloth, Thea Rolskov, and Astrid Sletten Rybner. 2023.
{``{DaDebias}/Genda-Lens.''} DaDebias.
\url{https://github.com/DaDebias/genda-lens}.

\leavevmode\vadjust pre{\hypertarget{ref-wang2017effectiveness}{}}%
Wang, Jason, Luis Perez, et al. 2017. {``The Effectiveness of Data
Augmentation in Image Classification Using Deep Learning.''}
\emph{Convolutional Neural Networks Vis. Recognit} 11 (2017): 1--8.

\leavevmode\vadjust pre{\hypertarget{ref-wei-zou-2019-eda}{}}%
Wei, Jason, and Kai Zou. 2019. {``{EDA}: Easy Data Augmentation
Techniques for Boosting Performance on Text Classification Tasks.''} In
\emph{Proceedings of the 2019 Conference on Empirical Methods in Natural
Language Processing and the 9th International Joint Conference on
Natural Language Processing (EMNLP-IJCNLP)}, edited by Kentaro Inui,
Jing Jiang, Vincent Ng, and Xiaojun Wan, 6382--88. Hong Kong, China:
Association for Computational Linguistics.
\url{https://doi.org/10.18653/v1/D19-1670}.

\end{CSLReferences}

\end{document}